\documentclass{article} 
\usepackage{iclr2020_conference,times}

\usepackage{amsmath,amsfonts,bm}

\def\eqref#1{equation~\ref{#1}}

\def\1{\bm{1}}

\DeclareMathAlphabet{\mathsfit}{\encodingdefault}{\sfdefault}{m}{sl}
\SetMathAlphabet{\mathsfit}{bold}{\encodingdefault}{\sfdefault}{bx}{n}

\usepackage{hyperref}
\usepackage{url}

\usepackage{graphicx}
\usepackage{makecell}
\usepackage{xcolor}
\usepackage{caption}
\usepackage{subcaption}
\usepackage{bm}
\usepackage{amsmath}

\title{EnsembleNet: End-to-End Optimization of Multi-headed Models}

\author{
Hanhan Li \qquad Joe Yue-Hei Ng \qquad Paul Natsev\\
\texttt{\{uniqueness,yhng,natsev\}@google.com} \\
Google AI\\
1600 Amphitheatre Parkway\\
Mountain View, CA 94043 \\
}

\iclrfinalcopy 
\begin{document}

\maketitle

\begin{abstract}
Ensembling is a universally useful approach to boost the performance of
machine learning models. However, individual models in an ensemble
were traditionally trained independently in separate stages without information access
about the overall ensemble. Many co-distillation approaches were proposed in order to treat model ensembling as first-class citizens.
In this paper, we reveal a deeper connection between ensembling and distillation, and come up with a simpler yet more effective co-distillation architecture.
On large-scale datasets including ImageNet, YouTube-8M, and Kinetics, 
we demonstrate a general procedure that can convert a single deep neural network
to a multi-headed model that has not only a smaller size but also better performance.
The model can be optimized end-to-end with our proposed co-distillation loss in a single stage without human intervention.
\end{abstract}

\vspace{-1em}
\section{Introduction}

In machine learning, ensemble methods combine multiple learning algorithms to obtain better predictive performance than could be obtained from any of the constituent learning algorithms alone \citep{opitz1999popular,polikar2006ensemble,rokach2010ensemble}. It is proven to be useful in a variety of domains including machine perception, natural language processing, user behavior prediction, and optimal control. Many top entries in Netflix Prize and Kaggle competitions are generated by a large ensemble of models. 

Traditionally, constituent models in an ensemble are trained independently in different stages and later on combined together. The process is laborious and requires manual interventions. Moreover, the constituent models are not ensemble aware and therefore not properly optimized. To jointly optimize the ensemble, one could naively train with a single loss on the final prediction, but the increased model size often leads to overfitting. Many recent works have studied different strategies to learn ensembles end-to-end by simultaneously optimizing multiple loss heads. In particular, \citet{lee2015m} studies multi-headed convolutional neural network (CNN) ensembles with shared base networks. Many recent approaches \citep{lan2018knowledge,song2018collaborative,lin2018nextvlad} also incorporate co-distillation losses.

In this paper, we extend the multi-headed approach and present the EnsembleNet architecture, where we use light-weight heads and a co-distillation loss which is simpler and more effective than previously used. We also reveal a similarity between conventional ensembling and distillation. Besides, previous ensembling approaches scale up the model size, but an EnsembleNet achieves much better performance than a single network with comparable model sizes in both training and inference, where the model size is measured in both the number of parameters and the number of FLOPs. We demonstrate this behavior extensively in a variety of large-scale vision datasets including ImageNet \citep{ILSVRC15}, YouTube-8M \citep{abu2016youtube}, and Kinetics \citep{kay2017kinetics}.

\section{Related Work}

There is extensive literature, going back decades, on ways to come up with and combine a group of strong and diverse models. Works related to ensembling can be broadly grouped into the following categories.

\textbf{Ensembling Theory.}\qquad Empirically, the prediction errors from individual models tend to cancel out when we ensemble them, and more diverse architectures tend to make better ensembles. This behavior can be explained from the point of view of Bayesian Model Combination \citep{domingos2000bayesian,minka2000bayesian,carroll2011turning,kim2012bayesian}. There are various theoretical models that estimate error bounds of specific ensemble formulations, given score distribution assumptions on the outputs and independence assumptions on the inputs \citep{kuncheva2014ensemble}.

\textbf{Ensembling Methods.}\qquad There are many works proposing specific ways to select candidate models and combine the predictions. A simple and popular ensembler just averages over the predictions of individual models. Other types of ensembling include greedy selection \citep{partalas2012study,li2012diversity,partalas2012study}, Mixture-of-Experts (MoE) fusion \citep{lan2018knowledge}, and sparsely gated MoE fusion \citep{shazeer2017outrageously,wang2018deep}. One may optionally incorporate the computation or memory cost into the optimization (e.g., the AdaNet algorithm in \citet{cortes2017adanet}). The ensembler may be trained either on the same data partition as the individual models, or on a separate partition. 

\textbf{Parameter Sharing.}\qquad Sharing a common base structure among multiple individual models may produce better ensembles, and this technique is used in \citet{lee2015m,lan2018knowledge}. Furthermore, hierarchical sharing may give additional performance boost \citep{song2018collaborative}.

\textbf{Ensemble-Aware Learning.}\qquad We would like to have a training strategy where individual models are aware of the ensembler during the optimization. A simple approach is to add a loss from the ensembler prediction. A related approach is co-distillation \citep{zhang2018deep,anil2018large,lan2018knowledge,song2018collaborative}, where constituent models are encouraged to learn from each other by regressing their predictions to the ensembler prediction. Our work provides a deeper insight on these approaches.

Also related to our work is shortcut auxiliary classifiers \citep{szegedy2015going,lee2015deeply}, which are used during training and discarded during inference. An EnsembleNet treats all individual models on equal footing and we don't have to tune their weights in loss. Nevertheless, one may also add shortcut auxiliary classifiers to an EnsembleNet as well.

\section{Approach}

\begin{figure}[ht]
    \centering
    \includegraphics[width=0.6\textwidth]{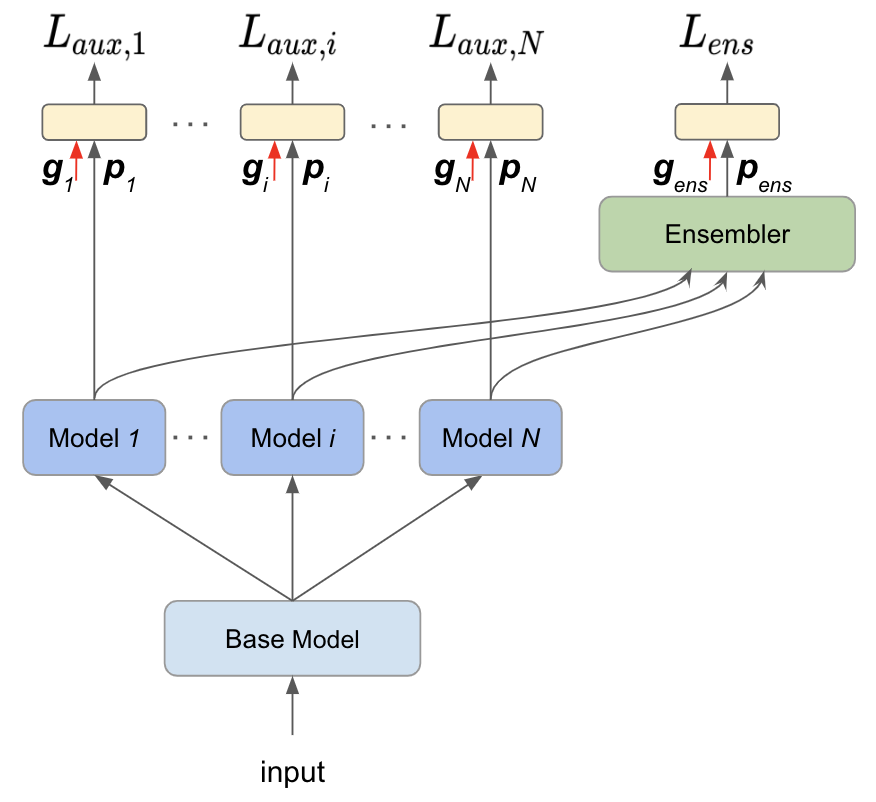}
    \caption{A multi-headed network with $N$ branches.}
    \label{fig:arch}
\end{figure}

We use a multi-headed network (Figure \ref{fig:arch}) and the output of each branch is an auxiliary prediction $\bm{p}_i$. As observed by \citet{lee2015m}, properly sharing the base network not only reduces computation resources but also increases model accuracy. An ensembler takes in all $N$ auxiliary predictions and outputs the final prediction $\bm{p}_{ens}$, which is used for inference. In training, a loss is computed for each prediction head and the final loss $L$ is the sum of all losses:
\begin{equation}
    L \equiv \sum_{i=1}^{N}L_{aux,i} +  L_{ens}
    \label{eqa:loss}
\end{equation}

Let $l(\bm{g}, \bm{p})$ be any loss function for measuring the discrepancy between a ground truth $\bm{g}$ and a prediction $\bm{p}$. We compare two ways of constructing $L_{aux,i}$ and $L_{ens}$.

The first one is the \textit{ensembling loss structure}.
\begin{equation}
\begin{cases}
    L_{aux,i}\equiv (1-\lambda)\cdot l(\bm{g}, \bm{p}_i) \\
    L_{ens} \equiv N\lambda \cdot l(\bm{g}, \bm{p}_{ens})
\end{cases}
\label{eqa:ensembling_loss}
\end{equation}
where we compute the loss for each prediction head against the same ground truth $\bm{g}$ for this task, and $\lambda$ is a scalar hyper-parameter that needs to be tuned. The rationale behind the coefficients is as follows. Suppose the simple average is used as the ensembler and the N branches have the same structure as well as initialization, then the loss is independent of $\lambda$. 

The second one is the \textit{co-distillation loss structure}.
\begin{equation}
\begin{cases}
    L_{aux,i}  \equiv \mu \cdot l(\bm{p}_{ens}, \bm{p}_i) \\
    L_{ens}    \equiv N \cdot l(\bm{g}, \bm{p}_{ens})
\end{cases}
\label{eqa:distillation_loss}
\end{equation}
where we compute the loss for each auxiliary prediction head against the ensembler prediction instead of the ground truth, and $\mu$ is a scalar hyper-parameter that needs to be tuned.

For the ensembling loss structure, if $\lambda=1$, we are directly optimizing for the ensembler prediction. This naive approach usually leads to severe overfitting and bad generalization for strongly performing individual models, which are typically quite large.  If $\lambda=0$, it is equivalent to the conventional ensembling and the resulting ensembler is typically much better than individual models. The overfitting problem is relieved here because an auxiliary loss will not be influenced by other branches, but one downside is that the auxiliary loss heads are not ensemble-aware. One might expect that an optimal $\lambda$ should be somewhere between $0$ and $1$. However, for most of the strongly performing networks we experimented with on YouTube-8M and ImageNet, the optimal values for $\lambda$ are negative! Basically if we decrease $\lambda$ from $1$ all the way to some negative number (e.g., $-1.5$), the performance of the resulting model decreases on the train set and increases on the holdout set, and the gap between them becomes smaller.

This observation is counter-intuitive, but we prove in Appendix \ref{subsec:proof} that if the ensembler does the simple averaging and $l$ is the $L_2$ loss, Equation \ref{eqa:ensembling_loss} and Equation \ref{eqa:distillation_loss} produce exactly the same loss if $\mu=1-\lambda$. The co-distillation term is an intuitive regularizer that encourages the individual model predictions to agree with each other, and choosing a negative $\lambda$ is nothing more than applying a strong distillation. A negative ensembler loss, which seemingly regresses the final prediction away from the ground truth, in fact helps regularize the models due to the presence of auxiliary losses. For other types of ensembler and other types of loss functions, we should still expect the effects from the two loss structures to be similar. In our experiments, we always use the simple average ensembler and the cross entropy loss function, with gradient stopped on $\bm{p}_{ens}$ in $l(\bm{p}_{ens}, \bm{p}_i)$. We found that the co-distillation loss with the best $\mu$ always slightly outperforms the ensembling loss with the best $\lambda$, and combining the two loss structures doesn't provide additional enhancement. This also means that co-distillation is preferable over the conventional ensembling.

Unlike previous end-to-end ensembling approaches  \citep{zhang2018deep,anil2018large,lan2018knowledge,song2018collaborative} that mix up the two loss structures by using both $l(\bm{g}, \bm{p}_i)$ and $l(\bm{p}_{ens}, \bm{p}_i)$, our EnsembleNet only applies the co-distillation loss (i.e., Equation \ref{eqa:distillation_loss}), which is structurally simpler and performs better.

Although we take simple average as the ensembler in our experiments, both loss structures can be generalized for any differentiable parametric ensembler, like an MoE model as in \citet{lan2018knowledge}, potentially with better results. In this case, we should also remove the coefficient $N\lambda$ for $L_{ens}$ in Equation \ref{eqa:ensembling_loss} and add it as a gradient multiplier to the input of the ensembler instead.

We invoke the following procedure to construct an EnsembleNet from a general deep neural network, where the EnsembleNet has both smaller size and better performance. We first take the upper half of a deep neural network (for example, a ResNet \citep{he2016deep} has 4 blocks and we take the upper 2 blocks), and shrink the width of the network (for example, the number of channels in a CNN layer) to reduce its size by more than a half (typically our shrinking ratio is about $1.5$). Then we duplicate this block once to build an EnsembleNet with two heads. Due to randomness in the initialization, the two branches with the same architecture will learn differently and become complimentary. The performance gain is robust against the exact layer from which we fork the network, and typically we choose some middle layers. Using different architectures for the two branches may yield even more complementary models and better results, but for simplicity, we choose the same architecture for them in the current experiments. We also found that using more than two branches gives at most marginally better performance, and results were omitted in the paper.  

\section{Experiments}
This section presents our detailed experiments on ImageNet \citep{ILSVRC15}, YouTube-8M \citep{abu2016youtube}, and Kinetics \citep{kay2017kinetics}. The reported accuracy metrics are computed using the sample mean of about 3 runs. The uncertainty of the sample mean, $\bar{x}$, for $x_1, x_2, ..., x_N$ is estimated with $\sqrt{\sum_{i=1}^{N}(x_i-\bar{x})^2/[N(N-1)]}$.

We measure both the number of model parameters and the FLOP counts for a model to report sizes. The FLOP count is computed with \texttt{tensorflow.profiler}, which treats multiplication and addition as two separate ops. We fold batch norm operations during inference when possible.

\subsection{ImageNet Classification}
\label{sec:imagenet}
ImageNet \citep{ILSVRC15} is a large scale quality-controlled and human-annotated image dataset. We use the ILSVRC2012 classification dataset which consists of 1.2 million training images with 1000 classes. Following the standard practice, the top-1 and top-5 accuracy on the validation set are reported.

\begin{figure}[ht]
    \centering
    \includegraphics[width=0.7\textwidth]{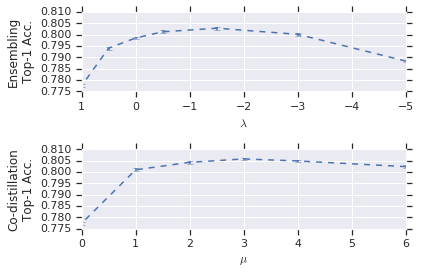}
    \caption{The validation set top-1 accuracies of multi-headed ResNet-152 with the ensembling loss (top) and the co-distillation loss (bottom). Detailed settings are specified in Table \ref{tab:imagenet_resnet}.}
    \label{fig:lambda_sweep}
\end{figure}

Table \ref{tab:imagenet_resnet} shows the performance and architecture comparisons between the baseline ResNet-152 \citep{he2016deep} and its multi-headed versions. Figure \ref{fig:lambda_sweep} plots the validation set top-1 accuracies for the ensembling loss with different $\lambda$ and the co-distillation loss with different $\mu$. ResNet-152 has 4 blocks. For the multi-headed variants, we leave the lower two blocks as they are, and construct two copies of the upper two layers as two branches. Each branch will have the bottleneck depth dimensions reduced, and the depth dimensions of the expanding $1\times1$ convolutions will be reduced proportionally. The width config in the table specifies the bottleneck dimensions for the 4 blocks.

The top-1 accuracy of our baseline ResNet-152 is $78.54\%$, which is higher than $77\%$ reported in the original implementation \citep{he2016identity}. The performance of the ensembling loss with $\lambda=1$ is lower than the baseline, showing that naively optimizing for the final prediction is not a good approach. The model with $\lambda=0.5$ and $\lambda=0$ gives $0.86\%$ and $1.31\%$ top-1 accuracy gain over the baseline respectively despite being slightly smaller. By decreasing $\lambda$ from $0$ to $-1.5$, we get an additional $0.43\%$ gain in top-1 accuracy, indicating a properly tuned negative $\lambda$ does reduce overfitting. An even more negative $\lambda$ degrades performance. For the co-distillation loss, the performance improves as $\mu$ increases from $0$ to around $3$ and slightly degrades when $\mu$ increases further. Please note that the top-1 accuracy with $\mu=3$ is $0.30\%$ higher than that with $\lambda=-1.5$.

Table \ref{tab:imagenet_seresnet} shows comparisons between the Squeeze-Excitation ResNet-152 \citep{hu2018squeeze} and its multi-headed versions. We set the reduction ratio to $16$ and use batch normalization in the Squeeze-Excitation layers. The top-1 accuracy of our baseline SE-ResNet-152 is $78.85\%$, which is higher than $78.43\%$ reported in \citet{hu2018squeeze}. The ensembling loss with $\lambda=0$ gives $1.91\%$ gain in top-1 accuracy, and the same loss with $\lambda=-1.5$ gives another $0.60\%$ gain. The co-distillation loss with $\mu=3$ enhances it by another $0.39\%$. All the observations are similar to those without Squeeze-Excitation.

Overall, the EnsembleNet (with $\mu=3$) offers a top-1 accuracy improvement of $\bm{2.04\%}$ for ResNet152 and $\bm{2.90\%}$ for SE-ResNet-152 on ImageNet. For comparisons, by keeping model sizes similar, the improvement is about $\bm{0.9\%}$ for co-distillation combined with MoE \citep{lan2018knowledge} and $\bm{1.0\%}$ for sparsely gated Deep MoE \citep{wang2018deep}.

\begin{table}[ht]
  \caption{Comparison of baseline models and their multi-headed versions on the ImageNet dataset. A resolution of $224\times224$ is used for both training and evaluation. Performance is evaluated using a single crop. The uncertainties of the accuracy metrics are about $0.06\%$.}
  {\small
  
  \begin{subtable}{1.0\textwidth}
      \centering
      \caption{ResNet-152 based models.}
      \label{tab:imagenet_resnet}
      \begin{tabular}{cccccc}
        \hline
        \makecell{ResNet-152 \\ based model}     & Width Config  & Top-1 Acc.     & Top-5 Acc.  & \#Params  & \#FLOPs \\
        \hline
        \hline
        Baseline & (64, 128, 256, 512) & 78.54\%  & 94.05\% & 60.1M & 21.8B \\
        \hline
        \makecell{$\lambda=1$ \\ or $\mu=0$}     &  \makecell{ base: (64, 128) \\ 2 branches: (176, 352)} & 77.74\% & 93.43\% & 58.2M & 21.0B \\
        \hline
        \makecell{$\lambda=0$}     &  \makecell{ base: (64, 128) \\ 2 branches: (176, 352)} & 79.85\% & 94.98\% & 58.2M & 21.0B \\
        \hline
        \makecell{$\lambda=-1.5$}  &  \makecell{ base: (64, 128) \\ 2 branches: (176, 352)} & 80.28\% & 95.27\% & 58.2M & 21.0B \\
        \hline
        \makecell{$\mu=3$}         &  \makecell{ base: (64, 128) \\ 2 branches: (176, 352)} & \textbf{80.58\%} & \textbf{95.38\%} & 58.2M & 21.0B \\
        \hline
      \end{tabular}
  \end{subtable}

  \bigskip
  \begin{subtable}{1.0\textwidth}
      \centering
      \caption{Squeeze-Excitation ResNet-152 based models.}
      \label{tab:imagenet_seresnet}
      \begin{tabular}{cccccc}
        \hline
        \makecell{SE-ResNet-152 \\ based model}     & Width Config  & Top-1 Acc.     & Top-5 Acc.  & \#Params  & \#FLOPs \\
        \hline
        \hline
        Baseline & (64, 128, 256, 512) & 78.85\%  & 94.25\% & 66.7M & 21.9B \\
        \hline
        \makecell{$\lambda=1$ \\ or $\mu=0$}     &  \makecell{ base: (64, 128) \\ 2 branches: (176, 352)} & 78.83\% & 93.75\% & 64.5M & 21.1B \\
        \hline
        \makecell{$\lambda=0$}     &  \makecell{ base: (64, 128) \\ 2 branches: (176, 352)} & 80.76\% & 95.33\% & 64.5M & 21.1B \\
        \hline
        \makecell{$\lambda=-1.5$}  &  \makecell{ base: (64, 128) \\ 2 branches: (176, 352)} & 81.36\% & 95.67\% & 64.5M & 21.1B \\
        \hline
        \makecell{$\mu=3$}         &  \makecell{ base: (64, 128) \\ 2 branches: (176, 352)} & \textbf{81.75\%} & \textbf{95.82\%} & 64.5M & 21.1B \\
        \hline
      \end{tabular}
  \end{subtable}
  }
\end{table}

\textbf{Implementation Details.}
The original ResNet architecture without pre-activation is used as the backbone for all experiments in this section. However, we made two slight modifications. First, for memory efficiency, we subsample the output activations in the last residual unit of each block, instead of subsampling the input activations in the first residual unit of each block. Second, the rectified linear units of the bottleneck layers are capped at $6$.

We use color augmentation as in \citet{howard2013some} in addition to scale and aspect ratio augmentation as in \citet{szegedy2015going}. Label smoothing is set to 0.1, and small l2 regularization is applied on all weights and batch norm $\gamma$ parameters. All convolutional weights are initialized using the same normal distribution with standard deviation of $0.03$. We use 128 TPU v3 cores (8x8 configuration) with a batch size of 32 in each core. The models are trained with the Momentum optimizer, where the learning rate starts at 0.01 and decays by 0.2 every 60 epochs.

\subsection{The YouTube-8M Video Classification}
YouTube-8M \citep{abu2016youtube} is a large-scale labeled video dataset that consists of features from millions of YouTube videos with high-quality machine-generated annotations. We use the 2018 version for the experiments where can compare with the best performing models in the Kaggle competition on this dataset. This version has about 6 million videos with a diverse vocabulary of 3862 audio-visual entities. 1024-dimensional visual features and 128-dimensional audio features at 1 frame per second are extracted from bottleneck layers of pre-trained deep neural networks and are provided as input features for this dataset.

Following \citet{abu2016youtube}, we measure the performance of our models in both global average precision (GAP) and mean average precision (mAP), with the number of predicted entities per video capped at 20. GAP is the area under curve for predictions across all video-entity pairs. For mAP, we first compute the area under curve for each entity across all videos, and then take the average across all entities. Following the standard of many YouTube-8M Kaggle participants, we train our models on the union of the train set and $90\%$ of the validation set and evaluate them on the remaining $10\%$ of the validation set. 

\begin{figure}[ht]
    \vspace{-.5em}
  \begin{center}
    \includegraphics[width=0.5\textwidth]{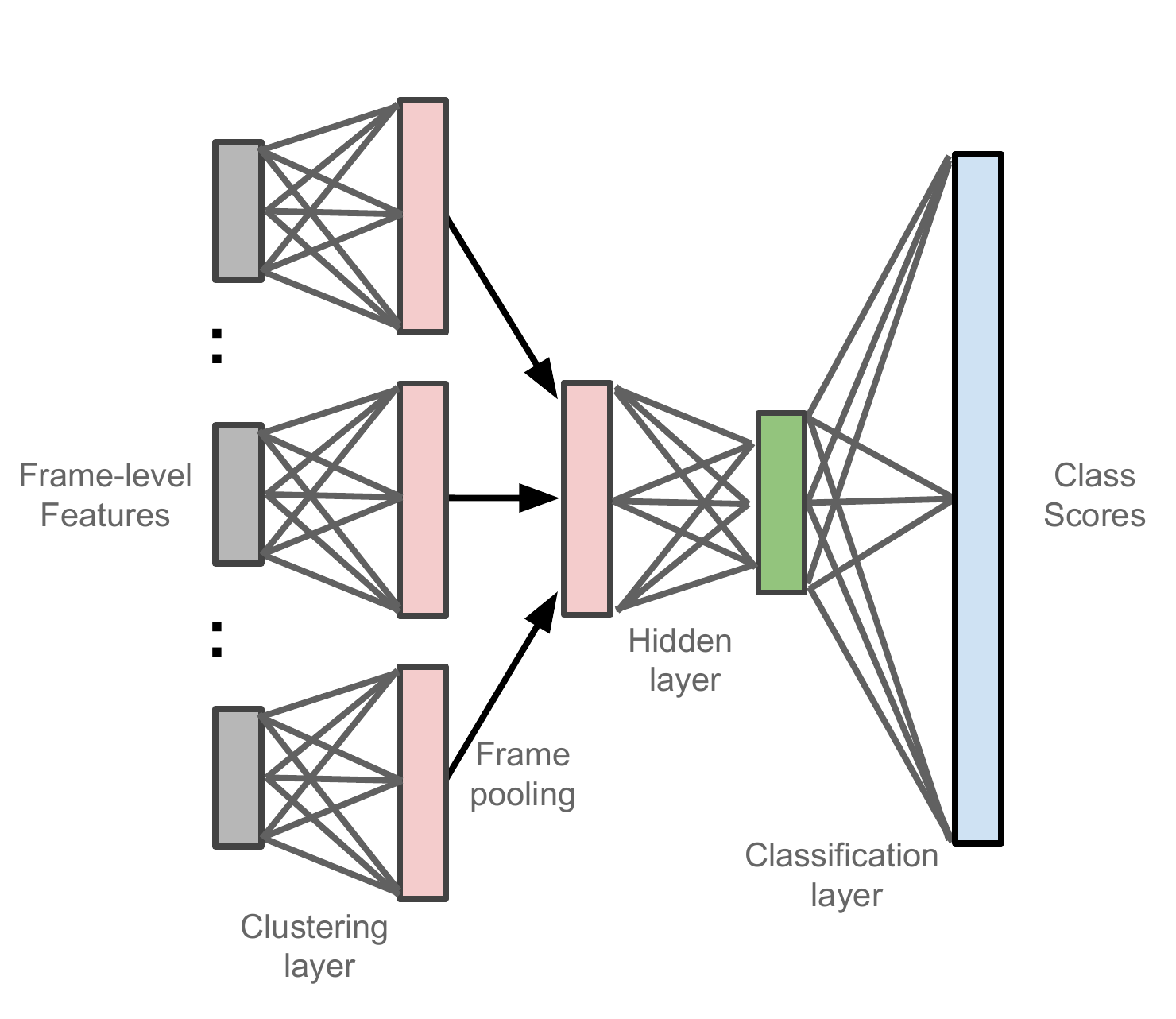}
  \end{center}
  \caption{\small The architecture of the deep bag of frame (DBoF) model for YouTube-8M.}
  \label{fig:dbof}
\end{figure}

Our baseline model is a variant of deep bag of frame (DBoF) model \citep{abu2016youtube} that incorporates context gates \citep{miech2017learnable}. The architecture of the model is shown in Figure \ref{fig:dbof}. We first pass the features of each frame through a fully-connected clustering layer to get the cluster representation, followed by a context-gating layer, and then a feature-wise weighted average pooling is used to extract a single compact representation. Afterward, we use a fully-connected hidden layer, followed by an MoE \citep{jordan1994hierarchical} classification layer to compute the final class scores. The classification layer uses one MoE for each class independently, and each MoE consists of several logistic experts.

The feature-wise weighted average frame pooling is defined as
\begin{equation}
    f(\{x_i|i=1, 2,...n\}) = \frac{\sum_{i=1}^{n}|x_i|x_i}{\sum_{i=1}^{n}|x_i|}
    \label{eqa:swap}
\end{equation}
where ${x_i}$ is the feature unit for the $i^{th}$ frame and $n$ is the number of frames. The intuition behind this pooling is that we would like to up-weight the features with large values so they don't get washed out in pooling. We refer to this non-parametric pooling method as Self-Weighted Average Pooling (SWAP).

The context gate \citep{miech2017learnable}, which is a multiplicative layer with a skip connection, is added in two places. One is in between the clustering layer and the frame pooling, and the other is after the classification layer. We also used batch normalization \citep{ioffe2015batch} on input and after any fully connected layer in the model.

For the multi-headed version, we fork right after the frame pooling layer, and in each branch the cluster layer size and the number of mixtures are reduced compared with the baseline DBoF.

Table \ref{tab:yt8m} shows the performance and architecture comparisons between the baseline DBoF and its multi-headed versions. The width config specifies the number of neurons in the cluster and hidden layers, as well as the number of logistic experts in the MoE classification layer. Our baseline DBoF has a GAP of $87.93\%$. The NeXtVLAD model \citep{lin2018nextvlad}, which is the best performing single model in 2018 YouTube-8M Kaggle competition, has a similar GAP of $87.85\%$. Although our DBoF is not as parameter efficient as NeXtVLAD, it is structurally simpler and serves as a good baseline for emsembling architectures. Our findings are similar to those from the ImageNet experiments. The ensembling loss with $\lambda=0$ gives a sizable performance gain despite being smaller. Setting $\lambda=-0.5$ produces roughly the best performance, which has a small gain in GAP with the same mAP. The co-distillation loss with $\mu=2$ is slightly better in both GAP and mAP.

\begin{table}[ht]
  \caption{Comparisons baseline DBoF with its multi-headed versions on the YouTube-8M dataset. The uncertainties of the accuracy metrics are about $0.01\%$. (The run-to-run variation is very small because the dataset uses pre-extracted high-level features instead of pixels as input.)}
  \label{tab:yt8m}
  \centering
  {\small
  \begin{tabular}{cccccc}
    \hline
    \makecell{DBoF \\ based model}     & Width Config  & GAP    & mAP & \#Params  & \#FLOPs \\
    \hline
    \hline
    Baseline & \makecell{cluster-4096 \\ hidden-4096, MoE-5} & 87.93\%  & 59.65\% & 229M& 415M\\
    \hline
    \makecell{ $\lambda=0$}  & \makecell{base: cluster-4096 \\ 2 branches: (hidden-3000, MoE-3)} & 88.30\% & 60.06\% & 226M& 410M  \\
    \hline
    \makecell{ $\lambda=-0.5$}     & \makecell{base: cluster-4096 \\ 2 branches: (hidden-3000, MoE-3)} & 88.35\% & 60.06\% &226M& 410M  \\
    \hline
    \makecell{ $\mu=2$}     & \makecell{base: cluster-4096 \\ 2 branches: (hidden-3000, MoE-3)} & \textbf{88.36}\% & \textbf{60.07}\% &226M& 410M  \\
    \hline
  \end{tabular}
  }
\end{table}

\textbf{Implementation Details.}
The models are trained with randomly sampling 25 frames and evaluated with taking all frames in each video. We use 32 TPU v3 cores (4x4 configuration) with a batch size of 16 in each core. We use the Adam optimizer, where the learning rate starts at 0.005 and decays by 0.95 every 1500000 examples. 

\subsection{Kinetics Action Recognition}
The Kinetics-400 dataset \citep{kay2017kinetics} contains about 240,000 short video clips with 400 action classes. Some videos are deleted over time so the train and validation sets have slightly fewer videos compared to the original version. We use the dataset snapshot captured in May 2019. Following the standard practice, the top-1 and top-5 accuracy on the validation set are reported.

We experimented with the S3D-G \citep{xie2018rethinking} model and its multi-headed counterparts. S3D-G consists of separable spatio-temporal convolutions and feature gating, and it gives a good speed-accuracy trade-off.

The performance and architecture comparisons are shown in Table \ref{tab:s3dg}. Similar to the notation in Section \ref{sec:imagenet}, we specify the model configuration with 4 numbers indicating the number of channels in the feature map right after each spatial sub-sampling in the network. Despite video removal in the dataset, our S3D-G baseline model closely matches the accuracy of the originally reported numbers (74.6\% in ours vs 74.7\% in \citet{xie2018rethinking}). We use two branches as before, and the number of channels in each branch is reduced by a factor of about 1.5 to maintain a comparable number of parameters and FLOPs with the original model. The multi-headed versions give large improvement over the baseline S3D-G model similar to what we observed before. The performance of the ensembling loss is roughly the best with $\lambda = 0$ or $\lambda = -0.5$, and the performance of the co-distillation loss with $\mu=2$ is slightly better.

\begin{table}[ht]
  \caption{Comparisons baseline S3D-G with its multi-headed versions on the Kinetics dataset. Our models take 64 RGB frames with $224\times224$ spatial resolution as input. The uncertainties of the accuracy metrics are about $0.1\%$.}
  \label{tab:s3dg}
  \centering
  {\small
  \begin{tabular}{cccccc}
    \hline
    \makecell{S3D-G \\ based model}     & Width Config  & Top-1 Acc.    &  Top-5 Acc. & \#Params  & \#FLOPs \\
    \hline
    \hline
    Baseline & \makecell{(64, 192, 480, 832) } & 74.6\%  & 91.4\% & 9.7M & 129B \\
    \hline
    \makecell{$\lambda=0$}     & \makecell{base: (64, 192, 480) \\ 2 branches: (554)} & 75.7\% & 92.2\% & 9.7M & 129B   \\
    \hline
    \makecell{$\lambda=-0.5$}     & \makecell{base: (64, 192, 480) \\ 2 branches: (554)} & 75.8\% & 92.1\% & 9.7M & 129B   \\
    \hline
    \makecell{$\mu=2$}        & \makecell{base: (64, 192, 480) \\ 2 branches: (554)} & \textbf{75.9\%} & \textbf{92.3\%} & 9.7M & 129B   \\
    \hline
  \end{tabular}
  }
\end{table}

\textbf{Implementation Details.}
Following \citet{feichtenhofer2018slowfast}, our S3D-G models use random initialization instead of ImageNet pretraining with a longer training schedule of 196 epochs. We use 128 TPU v3 cores (8x8 configuration) with a batch size of 16 in each core. The models are trained with the Momentum optimizer, where the learning rate starts at 0.8 and gradually decays to 0 at the end of training with the half-cosine decay schedule \citep{loshchilov2016sgdr}. 
\section{Conclusions}
In this paper, we presented a multi-headed architecture to train model ensembles end-to-end in a single stage. Unlike many previous works \citep{zhang2018deep,anil2018large,lan2018knowledge,song2018collaborative} that mix up the ensembling loss and the co-distillation loss, we only apply the co-distillation loss, where there is no need to explicitly regress individual model predictions to the ground truth. The co-distillation loss is theoretically related to the ensembling loss, and co-distillation empirically performs slightly better. Besides, contrary to the conventional belief that ensembling gains performance at the cost of higher computational resources, we demonstrated on a variety of large scale image and video datasets that we can scale down the size of an ensembled model to that of a single model while still maintaining a large accuracy improvement. Therefore, our approach is valuable for developing model ensembles from the perspectives of both automation and performance. Finally, EnsembleNet provides guidance on how to incorporate multiple loss heads in neural architectural search, by which we may potentially discover even better models.



\appendix
\section{Appendix}
\subsection{The Connection Between Ensembling and Co-distillation.}
\label{subsec:proof}
In this section, we prove that for the simple average ensembler and the $L_2$ loss, the ensembling loss is equivalent to the co-distillation loss. As a corollary, the conventional ensembling, which sets $\lambda=0$, has the co-distillation effect in it as well.

Let $\langle\bm{p}\rangle$ be the simple average of all of the predictions $\bm{p}_i$. The ensembling loss can be written as
\begin{align*} 
      &(1-\lambda)\sum_{i=1}^{N}l(\bm{g}, \bm{p}_i) + N\lambda l(\bm{g}, \bm{p}_{ens})  \\
      =& (1-\lambda)\sum_{i=1}^{N}(\bm{p}_i-\bm{g})^2
         + N\lambda (\langle\bm{p}\rangle-\bm{g})^2  \\
      =& (1-\lambda)\sum_{i=1}^{N}[(\bm{p}_i-\langle\bm{p}\rangle)^2 + 2(\bm{p}_i-\langle\bm{p}\rangle)\cdot(\langle\bm{p}\rangle-\bm{g}) + (\langle\bm{p}\rangle-\bm{g})^2]
         + N\lambda (\langle\bm{p}\rangle-\bm{g})^2  \\
      =& (1-\lambda)\sum_{i=1}^{N}(\bm{p}_i-\langle\bm{p}\rangle)^2 + N(\langle\bm{p}\rangle-\bm{g})^2  \\
      =& (1-\lambda)\sum_{i=1}^{N}l(\bm{p}_{ens}, \bm{p}_i) + N l(\bm{g}, \bm{p}_{ens})
\end{align*} 
which is the co-distillation loss.
\end{document}